\documentclass[letterpaper, 10 pt, conference]{ieeeconf}  

\IEEEoverridecommandlockouts                              

\usepackage{graphicx}
\usepackage{amsmath}
\usepackage{amssymb}
\usepackage{booktabs}
\usepackage{xcolor}

\usepackage{tabularx}
\usepackage{multirow}

\author{
\shortstack{
Takuma Tsukakoshi$^{1}$, Elle Miller$^{2}$,
Tamon Miyake$^{1}$, Namiko Saito$^{3}$,\\
Yushi Wang$^{1}$, Shigeki Sugano$^{1}$,
Tetsuya Ogata$^{1}$, Sethu Vijayakumar$^{2}$\\[0.5em]
{\small
$^{1}$Waseda University \quad
$^{2}$University of Edinburgh \quad
$^{3}$Microsoft Research Asia
}
}
}

\title{\LARGE \bf
Geometry-Aware Reinforcement Learning for \\Close-Fitting Wearable Insertion
}

\begin{document}

\maketitle
\thispagestyle{empty}
\pagestyle{empty}

\begin{abstract}
    Robot-assisted dressing is particularly challenging due to narrow openings and articulated body parts. Even small errors in the relative pose between the body part and garment opening can cause snagging, missed insertion, or incomplete dressing. To address these challenges, we formulate dressing as a geometry-aware insertion problem that represents task-relevant spatial relationships between the body part and object opening, rather than relying on the full deformable-object state. Based on this formulation, we propose a Reinforcement Learning (RL) framework combining a staged curriculum, which progressively guides the policy from coarse alignment to insertion and final placement, with a finger-wise insertion soft gate that provides continuous feedback on individual finger insertion progress and helps suppress premature insertion. 
    We evaluate the framework on rigid and deformable bracelet dressing tasks through component ablations, systematic variations in the initial hand pose, and closed-loop execution of the learned policies on a real robotic system. On the post-training deformable evaluation, the full method reaches $84.7\pm 1.2\%$ success. The insertion gate accounts for the largest gain, and adding the curriculum on top of that gate reaches the top score. On a mannequin hand the same full policy succeeds on $15/20$ real-robot trials.
\end{abstract}

\section{INTRODUCTION}
Assistive dressing requires a robot to pass the opening of a wearable over a human limb while ensuring safety and precision~\cite{liang2025openrobocare}.
For loose garments, such as sleeves or jackets, dressing mainly involves reaching and guidance problems. In contrast, for close-fitting wearables, dressing is fundamentally a precise insertion problem: a body part must pass through a narrow, deformable opening, where small misalignments can collapse the opening, cause snagging, or result in incomplete insertion~\cite{Tsukakoshi2026close}.

For loose garments, existing dressing systems already solve several related tasks in both simulation and real-robot deployment.
Policies trained via real-world demonstrations can pull sleeves or jackets onto an arm when the opening is wide and the initial placement is constrained~\cite{zhang2022learning,zhu2024you}.
Those policies, however, inherit the demonstrator's skill and coverage, and the collected motions assume a wide opening.
In simulation, caregiving environments such as Assistive Gym and RCareWorld dress gowns and sleeves, typically on simplified body geometry~\cite{erickson2020assistive, ye2022rcareworld}.
Reinforcement learning (RL) policies can dress hospital-gown sleeves onto a human model with spherical hands~\cite{erickson2020assistive}, later extended to collaborative human--robot dressing~\cite{Clegg2020learning}.
DexGarmentLab constructed a simulated glove-dressing task in which both hands insert the index fingers, open the cuffs, and pull the gloves onto the hands; the sequence was completed with an attachment-block workaround~\cite{wang2025dexgarmentlab}. However, the small opening, the required index-finger insertion, and the subsequent thumb pinch were difficult under their FEM cloth model and dexterous-hand control.
Separately, learning-based cloth methods have achieved unfolding and related garment skills, including successful sim-to-real transfer~\cite{longhini2025unfolding,wang2025robot}.

While these methods have advanced loose-garment dressing and general
cloth manipulation, their results do not directly address the constraints of close-fitting wearables.
Close-fitting placement introduces three distinct challenges.
First, geometric tolerance is tight: unlike loose garments with openings
substantially larger than the target body part~\cite{zhang2022learning},
close-fitting wearables leave little clearance for relative pose error
during insertion.
Second, insertion is sequential and contact-rich: successful dressing requires a sequence of coordinated interactions between the wearable opening and the body part, where local snagging or misalignment can prevent subsequent progress.
Third, deformable-object state is difficult to represent consistently across simulation and the real world: a full garment mesh is difficult to reconstruct from camera observations, while simulated cloth states depend on imperfect models of thin-contact dynamics~\cite{longhini2025unfolding}.

We address these difficulties with three corresponding approaches.
To meet the tight geometric tolerances of close-fitting wearable dressing, we formulate the RL problem as \textit{geometry-conditioned insertion},
using bracelet dressing as a representative instance.
The policy conditions on proprioception and the relative pose between
the hand and the wearable opening.
To mitigate snagging and incomplete enclosure, we train an RL policy
with a staged curriculum that progresses from little-finger insertion
to thumb insertion and finally wrist enclosure, together with a soft
finger-wise insertion gate that modulates the wrist-stage reward based
on insertion progress.
To avoid both mesh reconstruction and mesh-dependent simulation, these geometric features are keypoints that can be measured from cameras, not a cloth mesh.

\begin{figure}[t]
    \vspace{0.2cm}
    \centering
    \includegraphics[width=0.98\columnwidth]{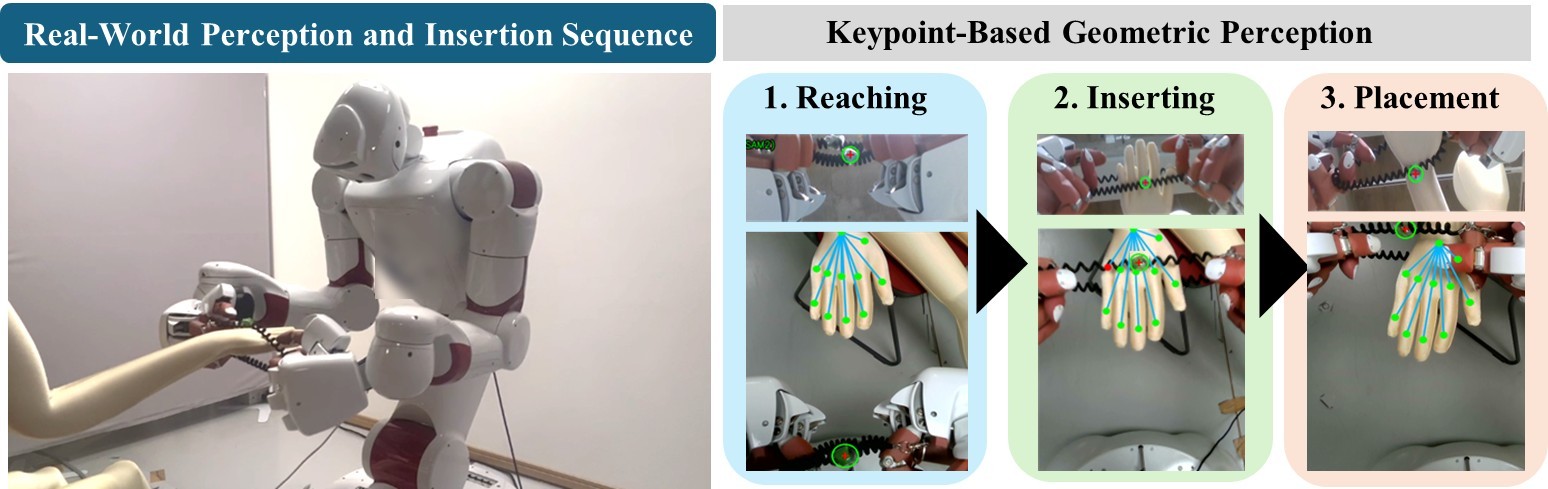}
    \caption{\textbf{Real-world closed-loop dressing.} Camera keypoints of the hand and opening guide reaching, insertion, and placement on a mannequin.}
    \label{fig:real}
        \vspace{-0.4cm}
\end{figure}

We evaluate the framework primarily on a deformable bracelet dressing task, using a rigid bracelet as a diagnostic baseline.
Ablations isolate the staged curriculum and the soft gate.
We also test robustness to moderate in-domain hand-pose variation and demonstrate the learned policy on a real robotic system (Fig.~\ref{fig:real}).
The contributions of this work are as follows:
\begin{itemize}


    \item \textbf{Geometric formulation of close-fitting dressing to aid transfer:}
    We represent the task using task-relevant relative geometry between the articulated body part and the wearable opening, rather than relying on the full deformable garment state, thereby maintaining consistency between simulation and real-world execution. 

    \item \textbf{Close-fitting curriculum and finger-wise insertion soft gate:}
    We guide learning from little-side alignment to thumb-side alignment and wrist insertion, and introduce a finger-wise soft insertion gate that modulates the wrist reward by containment within the object opening. This reduces premature failures such as finger snagging, opening collapse, and incomplete enclosure.

    \item \textbf{Zero-shot sim-to-real transfer of close-fitting dressing with deformable openings:}
    We reconstruct the simulation geometric observation from real-world cameras and deploy the learned policies in closed loop without additional real-world training, demonstrating transfer to deformable bracelet insertion.

\end{itemize}

\begin{figure*}[t]
    \vspace{0.2cm}
    \centering
    \includegraphics[width=0.98\textwidth,trim={0 0.85cm 0 0},clip]{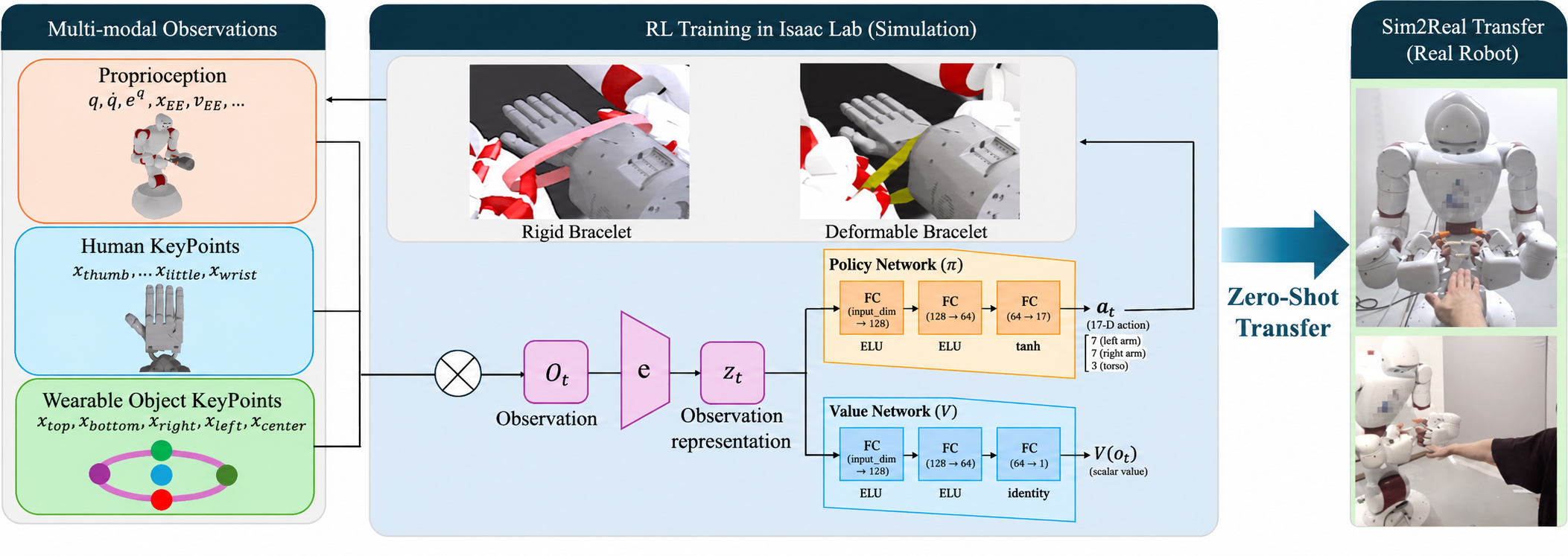}
    \caption{\textbf{Method overview.} The policy uses robot proprioception, user hand keypoints, and wearable-object keypoints to encode relative hand-object geometry and generate dressing actions. The primary evaluation is the \emph{Deformable Bracelet} task; a \emph{Rigid Bracelet} diagnostic isolates insertion when the opening cannot collapse.}
    \label{fig:model}
\end{figure*}

\section{Related Work}
\label{related work}
   	\subsection{Robot-assisted dressing.}
    Robot-assisted dressing has been widely studied as a challenging assistive manipulation problem that requires sustained physical interaction with the human body \cite{chen2026feeding}. 
    Several studies employed demonstration-driven visuomotor policy frameworks to learn dressing skills from human demonstrations and teleoperation data \cite{joshi2019framework,Tsukakoshi2026close,zhu2024you,pignat2017learning}. These approaches achieved stable garment manipulation and human-arm guidance under constrained settings. \cite{zhang2022learning} addresses sleeve-to-arm insertion by training grasping policies with supervised learning in a simulator, which is conceptually related but uses a wide garment opening and does not require sequential insertion of five articulated fingers. 
    
    Another popular approach is RL. Caregiving simulators such as RCareWorld \cite{ye2022rcareworld} and Assistive Gym \cite{erickson2020assistive} offer dressing environments for gowns and sleeves, using simplified body geometry. For instance, \cite{erickson2020assistive} simulates hospital-gown sleeve dressing over the arm. \cite{Clegg2020learning} proposes a co-optimization method for collaborative human-robot dressing, validated in Assistive Gym. More recently, DexGarmentLab \cite{wang2025dexgarmentlab} studied bimanual glove dressing, but relied on attachment blocks due to the difficulty of dexterous insertion under FEM simulation. Our setting differs from prior works in the combination of a fully articulated hand, a close-fitting bracelet opening, and a reward that gates wrist insertion on per-finger containment. The gated distance terms follow the same insertion-style structure as prior sleeve rewards \cite{erickson2020assistive}; the added mechanism is the finger-wise soft gate and the little-to-thumb staging.
    \cite{wang2023one} proposed a dressing policy that generalizes across diverse human arm poses and garments by leveraging partial point-cloud observations and policy distillation to integrate policies trained over different ranges of arm poses.

    
    \subsection{Garment representation and sim-to-real transfer.}
    Deformable manipulation has advanced as much through garment simulation as through learning algorithms. Particle- and FEM-based cloth simulators supply the interactions RL needs and have enabled cloth, rope, and garment skills \cite{lin2021softgym,wang2025robot,tsurumine2019deep,hoque2020visuospatial,wang2025goal,puthuveetil2022bodies,louis2026se3}, but they model poorly the regime dressing occupies: thin-shell self-contact, skin friction, and buckling of a stretched opening \cite{longhini2025unfolding,wang2025dexgarmentlab}. For dressing, the sim-to-real gap therefore sits in the garment dynamics, not in the arm. Demonstrations avoid that gap by being collected on hardware, but cover only the openings and body poses the demonstrator visited \cite{joshi2019framework,zhu2024you,pignat2017learning}; RL explores far more configurations, inside a simulator whose cloth model is the weak link. The decisive choice is thus not imitation versus RL but \emph{what the policy is told about the garment}, since that sets how much of the simulator it inherits. Dense states, such as meshes, particles, learned latents \cite{kotsovolis2024garment}, dynamics models \cite{chen2025graphgarment}, are defined by the cloth model and cannot be reconstructed from one camera view; partial point clouds \cite{wang2023one} and multimodal sensing \cite{zhang2023visual,sun2024force,hao2025force} relax these requirements at the cost of additional hardware; a few sparse task variables lie at the other end \cite{tamei2011reinforcement}. Geometry-aware graphs of object and actuator poses have been used for insertion and cloth manipulation~\cite{hoang2025geometryaware}. We take that sparse end: the relative hand--opening pose is measurable from cameras and independent of any cloth model. Geometry is not universally better---cloth tasks do transfer with denser state \cite{longhini2025unfolding}---but it suffices for bracelet insertion, where the difficulty is fine alignment over a long contact horizon rather than global garment shape \cite{Tsukakoshi2026close}.

\section{Method}
\label{sec:citations}

\subsection{Problem Setting}

We study geometry-conditioned bracelet insertion, in which an upper-body humanoid robot places a wearable object onto an articulated human hand (Fig.~\ref{fig:model}). 
Although the task involves contact-rich interactions and object deformation, the policy relies on task-relevant relative geometry between the hand and the wearable opening, rather than explicit contact or full deformable-object state.
Due to the lack of a compact state representation for deformable objects, we formulate the problem as a Partially Observable Markov Decision Process (POMDP).
The state space $\mathbf{s} \in \mathcal{S}$, action space $\mathbf{a} \in \mathcal{A}$, transition kernel $T : \mathcal{S} \times \mathcal{A} \times \mathcal{S} \rightarrow [0,1]$, reward function $R : \mathcal{S} \times \mathcal{A} \rightarrow \mathbb{R}$, and discount factor $\gamma \in (0,1)$ follow standard definitions. The true state $\mathbf{s}$ is unobservable; instead, the agent receives an observation $\mathbf{o} \in \mathcal{O}$ and acts according to a policy $\pi : \mathcal{O} \rightarrow \mathcal{P}(\mathcal{A})$ that parameterizes a distribution over actions, with the objective of finding the optimal policy $\pi^{\star} = \mathrm{arg\,max}_{\pi} \mathbb{E}\left[\sum_{t=1}^{\infty}\gamma^{t-1}R(\mathbf{s_t},\mathbf{a_t})\right]$. The hidden state includes garment deformation, contact, slipping, wrinkling, and snagging, none of which we assume to be directly measurable.

\textbf{Success criterion.}
We define success using a geometric containment test rather than
proximity alone (Figs.~\ref{fig:rew_set} and~\ref{fig:failures}).
For each finger $i$, we define the signed distance to the opening plane as
$d_i=p^{\mathrm{base}}_{i,x}-c^{\mathrm{opening}}_x$,
where positive and negative values correspond to the approach and
insertion sides of the opening, respectively.
A finger is successfully inserted when its base crosses the opening plane from $d_i>+3$ mm to $d_i<-3$ mm within the opening ellipse ($e_i\leq1$).
An episode is successful only when all five fingers satisfy this
insertion criterion and the wrist-to-opening distance is less than
$1$ cm.
Thus, placing the opening near the hand without enclosing the wrist
is not counted as success.

\subsection{Geometry-conditioned POMDP with staged exploration and gating}
\begin{figure*}[t]
    \vspace{0.2cm}

    \centering
    \includegraphics[width=0.8\linewidth,trim={0 0.45cm 0 0.5cm},clip]{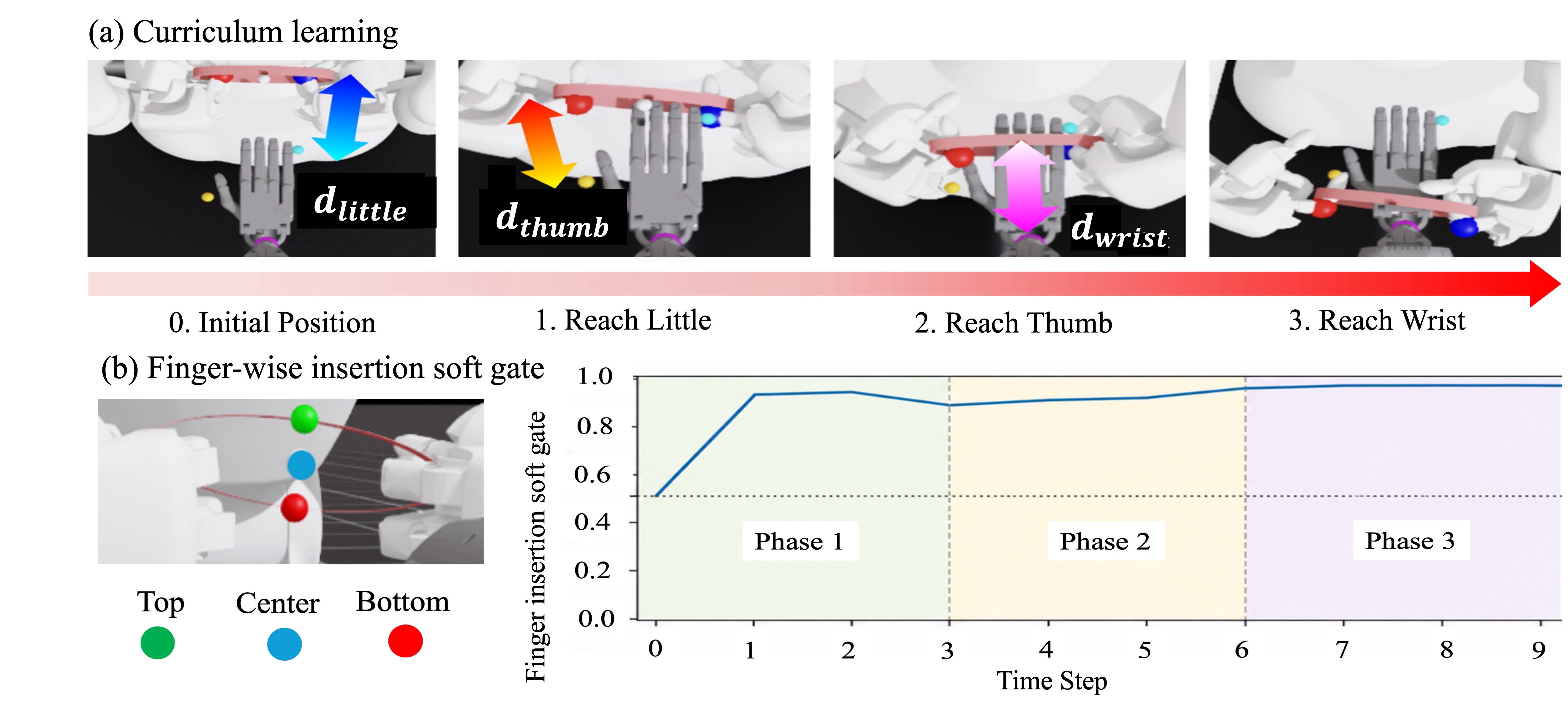}
    \caption{\textbf{Curriculum and the finger-wise soft gate.}
    (a)~Little-side, thumb-side, and wrist distances define the staged reward from coarse alignment to wrist insertion.
    (b)~The soft gate uses the top, center, and bottom opening keypoints and saturates as more fingertips enter the opening height band.}
    \label{fig:rew_set}
\end{figure*}

    \textbf{Geometry-aware observation space $\mathcal{O}$.}
    The policy observation $\mathbf{o}_t \in \mathcal{O}$ combines robot
    proprioception with task-relevant geometric features. The proprioceptive component
    $\mathbf{o}^{\mathrm{prop}}_t$ summarizes the state of the robot, including the joint states of the arm and torso, the poses of the end-effector and the previous action. The geometric component
    $\mathbf{o}^{\mathrm{gt}}_t$ encodes compact relative relationships among the
    robot end-effectors, human target keypoints, and the corresponding keypoints on
    the wearable object. As illustrated in Fig.~\ref{fig:rew_set}(a), these keypoint
    correspondences define the task-relevant spatial relationships required for
    alignment and insertion. This representation provides the policy with local
    geometric cues for contact-rich assistive dressing without completely modeling
    the garment state. Table~\ref{tab:obs} lists the observation components used by the policy. The vector $g^{\mathrm{finger}}_{\mathrm{soft}} \in \mathbb{R}^{5}$ concatenates a soft insertion score for each finger, given by a sigmoid of that fingertip's signed margin to the opening height band $[h_{\mathrm{bottom}}, h_{\mathrm{top}}]$.

    \textbf{Action space $\mathcal{A}$.}
    The policy action $\mathbf{a}_t \in \mathbb{R}^{17}$ consists of joint-position
    commands for the 14 arm DoFs and the 3 torso DoFs.
    The torso is included so that the bimanual grasp can be repositioned relative
    to the human hand: a locked torso fixes the shoulder origins, which restricts
    the set of opening poses that both arms can reach when the hand is offset.
    The head and the grasping fingers remain excluded from policy control and stay
    in a fixed configuration.
    We freeze the grasp at the start of each episode to isolate dressing from
    in-hand regrasping, matching the real-robot pinch.

\begin{table}[t]

\centering
\caption{Observation space. The policy input $\mathbf{o}_t$ consists of robot proprioceptive and task-relevant geometric features.}
\label{tab:obs}
\scriptsize
\setlength{\tabcolsep}{5pt}
\begin{tabular}{@{}cccc@{}}
\toprule
Type & Symbol & Description & Dim. \\
\midrule
\multirow{8}{*}{$\mathbf{o}^{\mathrm{prop}}_t$}
& $\bar{\mathbf{q}}_t$ & normalized arm and torso joint positions & $17$ \\
& $\bar{\dot{\mathbf{q}}}_t$ & normalized arm and torso joint velocities & $17$ \\
& $\mathbf{e}^{q}_t$ & arm and torso joint position errors & $17$ \\
& $\mathbf{x}^{R}_{\mathrm{EE},t}$ & right end-effector position & $3$ \\
& $\mathbf{r}^{R}_{\mathrm{EE},t}$ & right end-effector orientation & $4$ \\
& $\mathbf{x}^{L}_{\mathrm{EE},t}$ & left end-effector position & $3$ \\
& $\mathbf{r}^{L}_{\mathrm{EE},t}$ & left end-effector orientation & $4$ \\
& $\mathbf{a}_{t-1}$ & previous action & $17$ \\
\midrule
\multirow{9}{*}{$\mathbf{o}^{\mathrm{gt}}_t$}
& $\Delta \mathbf{x}^{\mathrm{EE}}_t$ & relative EE position & $3$ \\
& $d^{\mathrm{EE}}_t$ & EE--EE distance & $1$ \\
& $\Delta \mathbf{x}^{R,\mathrm{thumb}}_t$ & right EE to thumb & $3$ \\
& $d^{R,\mathrm{thumb}}_t$ & right EE--thumb distance & $1$ \\
& $\Delta \mathbf{x}^{L,\mathrm{little}}_t$ & left EE to little finger & $3$ \\
& $d^{L,\mathrm{little}}_t$ & left EE--little distance & $1$ \\
& $\Delta \mathbf{x}^{\mathrm{wrist}}_t$ & relative wrist position & $3$ \\
& $d^{\mathrm{wrist}}_t$ & wrist-associated distance & $1$ \\
& $g^{\mathrm{finger}}_{\mathrm{soft}}$ & per-finger soft insertion scores & $5$ \\
\bottomrule
\end{tabular}
\end{table}

\textbf{Curriculum-based reward function $\mathcal{R}$.}
The reward is designed to guide the wearable object from coarse little- and
thumb-side alignment to wrist-level insertion, while promoting finger-wise progression through the wearable opening toward a geometrically valid insertion configuration. The task is decomposed into three geometry-conditioned subtasks: little-side alignment, thumb-side alignment,
and wrist insertion. Each subtask is represented by a gated distance-based reward
term, weighted by a fixed scale. Although the scales remain fixed throughout
training, the gates determine when each subtask can contribute to the reward.
This induces an implicit curriculum in which the policy first learns valid finger-side alignment and then prioritizes wrist-level insertion once the finger reaches a valid insertion pose within the object opening.
For each reward component $k$, we define the gated distance reward as
\begin{equation}
    r_k = a_k r(D_k;\sigma_k),
\end{equation}
where $D_k$ is the task-specific distance, $a_k$ is the corresponding activation
gate, and
\begin{equation}
    r(D_k;\sigma_k)=1-\tanh(D_k/\sigma_k)
\end{equation}
maps the Euclidean distance to a reward in $(0,1]$. 
\begin{table}[t]
    \centering
    \caption{Reward components and activation gates for wearable-object dressing.}
    \label{tab:reward_terms}
    \scriptsize
    \setlength{\tabcolsep}{6pt}
    \begin{tabular}{@{}ccccc@{}}
    \toprule
    $r_k$ & $D_k$ & $\sigma_k$ & $\lambda_k$ & $a_k$\\
    \midrule
    $r_{\mathrm{ee\_little}}$
    & $\|\mathbf{x}_{\mathrm{left\_ee}}-\mathbf{x}_{\mathrm{little}}\|$
    & $0.05$ & $10.0$ & $c_{\mathrm{ee}}\,c_{\mathrm{little}}$ \\
    $r_{\mathrm{ee\_thumb}}$
    & $\|\mathbf{x}_{\mathrm{right\_ee}}-\mathbf{x}_{\mathrm{thumb}}\|$
    & $0.10$ & $20.0$ & $c_{\mathrm{ee}}\,c_{\mathrm{thumb}}$ \\
    $r_{\mathrm{wrist\_opening}}$
    & $\|\mathbf{x}_{\mathrm{wrist}}-\mathbf{x}_{\mathrm{opening}}\|$
    & $0.14$ & $200.0$ & $c_{\mathrm{ee}}\,g_{\mathrm{fingers}}$ \\
    \bottomrule
    \end{tabular}
\end{table}
The total reward is then defined as
\begin{equation}
    \mathcal{R}
    =
    \lambda_{\mathrm{little}}
    r_{\mathrm{ee\_little}}
    +
    \lambda_{\mathrm{thumb}}
    r_{\mathrm{ee\_thumb}}
    +
    \lambda_{\mathrm{wrist}}
    r_{\mathrm{wrist\_opening}},
\end{equation}
where $\lambda_{\mathrm{little}}$, $\lambda_{\mathrm{thumb}}$, and
$\lambda_{\mathrm{wrist}}$ are fixed reward scales. We set
$\lambda_{\mathrm{little}}=10.0$, $\lambda_{\mathrm{thumb}}=20.0$, and
$\lambda_{\mathrm{wrist}}=200.0$ to prioritize task completion. The distance
definitions, scales, and activation gates are summarized in
Table~\ref{tab:reward_terms}.
To ensure that the policy is rewarded only for geometrically valid progress, the
alignment rewards are activated using hand-object geometric gates.  Here, $c_{\mathrm{ee}}=\mathbb{I}[d_{\mathrm{ee}} < 0.3]$ is a proximity gate
that activates the reward only when the end-effector is close to the hand-object
interaction region. The little-side
and thumb-side alignment rewards are enabled only when the corresponding finger is
located inside the height band of the wearable opening:
\begin{align}
    c_{\mathrm{little}}
    &=
    \mathbb{I}
    \bigl[
    h_{\mathrm{bottom}} < h_{\mathrm{little}} < h_{\mathrm{top}}
    \bigr],
    \\
    c_{\mathrm{thumb}}
    &=
    \mathbb{I}
    \bigl[
    h_{\mathrm{bottom}} < h_{\mathrm{thumb}} < h_{\mathrm{top}}
    \bigr].
\end{align}
Here, $h_{\mathrm{top}}$ and $h_{\mathrm{bottom}}$ denote the height coordinates
of the top and bottom keypoints of the wearable opening, and $h_i$ denotes the height coordinate of finger $i$. 

\textbf{Finger-wise insertion soft gate}.
For wrist insertion, we introduce a finger-wise insertion soft gate that softly
quantifies whether individual fingers are located within the opening height
band. For each finger $i \in \mathcal{F}$, we compute the signed margin to the
nearest opening boundary:
\begin{equation}
    m_i =
    \min \left(
    h_i - h_{\mathrm{bottom}},
    h_{\mathrm{top}} - h_i
    \right).
\end{equation}
A positive margin indicates that the finger is inside the opening height band,
while a negative margin indicates that it is outside. We convert these margins
into a differentiable gate by averaging their sigmoid responses:
\begin{equation}
    g_{\mathrm{fingers}}
    =
    \frac{1}{|\mathcal{F}|}
    \sum_{i \in \mathcal{F}}
    \operatorname{sigmoid}
    \left(
    \frac{m_i}{\tau}
    \right),
\end{equation}
where $\tau=0.01$ is the temperature parameter, chosen so that the gate saturates once a fingertip is clearly inside the band. We compare this soft gate to a hard indicator in Section~\ref{sec:result}.

     The fixed scales set the relative priorities of the subtasks: little-side alignment provides initial guidance, thumb-side alignment refines the opening pose, and wrist insertion receives the largest weight to emphasize the final stage of dressing. The gates control when each reward term becomes active, preventing the high wrist-insertion weight from encouraging premature insertion before the fingers are properly positioned within the wearable opening. Together, these mechanisms induce an implicit curriculum without changing the weights during training, progressing from little-side alignment to thumb-side alignment and finally to wrist-level insertion.

    For deformable objects, the opening keypoints are not assumed to be motion-capture markers. We select four rim vertices (top, bottom, left, right) and a center point from the rest-shape mesh using a radial PCA band around the principal axis, then track those fixed vertex indices as the mesh deforms. This keeps the target labels temporally consistent without re-segmenting the garment at every step. The same geometric targets are instantiated on the real bracelet from SAM2 tracks of the top and bottom opening markers and their midpoint.
         
    \textbf{Reset function.}
    To improve sample efficiency, episodes terminate at the time limit or in unrecoverable states: when a grasp point moves more than $0.30\,\mathrm{m}$ from its end-effector, the end-effectors separate by more than $0.50\,\mathrm{m}$, or the wearable detaches and falls below $0.40\,\mathrm{m}$.

\section{Experimental setup}

\begin{figure*}[t]
    \vspace{0.2cm}

    \centering
    \includegraphics[width=0.9\textwidth]{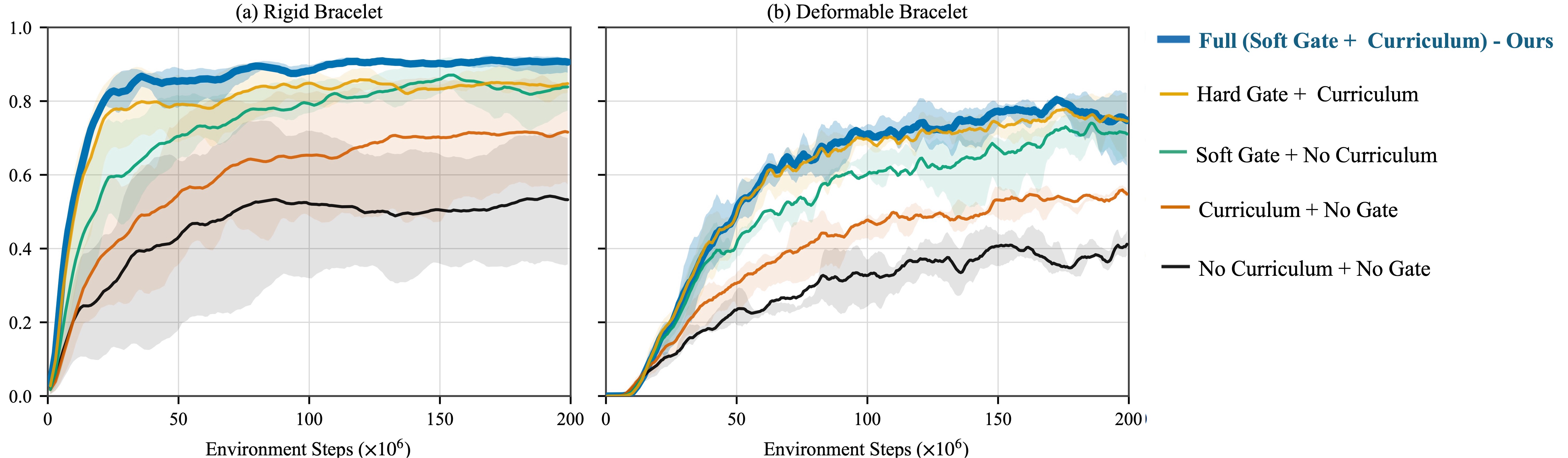}
    \caption{\textbf{Mean evaluation success rate during training for the rigid and deformable bracelet dressing tasks.}
    Solid lines are the three-seed mean; shaded regions are the min--max range.
    These curves are not the post-training $50$-episode scores in Table~\ref{tab:ablation}.
    Full (soft gate + curriculum) is highest on both panels.
    The insertion gate accounts for the large gap; the deformable ceiling is lower than the rigid diagnostic because the opening can collapse.}
    \label{fig:success_rate}
\end{figure*}

\textbf{RL.} We used a customized implementation
of Proximal Policy Optimization (PPO) \cite{schulman_proximal_2017} from SKRL~\cite{serrano2023skrl} to incorporate observation stacking and separated environments for continuous agent evaluation \cite{miller2025tactilerl}. The agent comprised an early-fusion MLP encoder ($512$--$256$), a policy head ($128$--$64$), and a value head ($128$--$64$), all with ELU hidden activations. Across experiments, we kept the PPO hyperparameters fixed except for the learning rate: $32$ rollouts, $8$ epochs, $16$ mini-batches, $\gamma=0.99$, $\lambda=0.95$, ratio clip $0.2$, and learning rates $1.0\times 10^{-5}$ (\emph{Rigid Bracelet}) and $1.0\times 10^{-6}$ (\emph{Deformable Bracelet}). We used $4096$ / $200$ training/evaluation environments for \emph{Rigid Bracelet} and $2500$ / $200$ for \emph{Deformable Bracelet}. All compared policies are trained for $200$\,M steps (Fig.~\ref{fig:success_rate}).

\textbf{Simulation.} We evaluate our method in two custom Isaac Lab environments~\cite{isaaclab}. \emph{Deformable Bracelet} is the primary task, with an opening of approximately $0.10\,\mathrm{m} \times 0.060\,\mathrm{m}$. \emph{Rigid Bracelet} is a diagnostic with a larger, non-collapsible opening of approximately $0.25\,\mathrm{m} \times 0.075\,\mathrm{m}$, used to test whether the same geometric rewards suffice when deformation was removed. The wearable object was pre-grasped, and the policy commanded the $14$ arm joints and the $3$ torso joints. The physics timestep was $0.002\,\mathrm{s}$ with decimation $10$ (control interval $0.02\,\mathrm{s}$).

\textbf{Compute.} Experiments were run on a single NVIDIA GeForce RTX 5090 GPU. During training, Isaac Lab used approximately 31\,GB of VRAM. A rigid-bracelet run of $200$\,M steps took approximately $15$ hours. The long training time was due to the fine-grained physics step required to reduce penetration and tunneling artifacts. Deformable-bracelet training used the same step budget and is slower per step under FEM contact.

\begin{figure}[t]
    \vspace{0.2cm}
    \centering
    \includegraphics[width=0.98\linewidth]{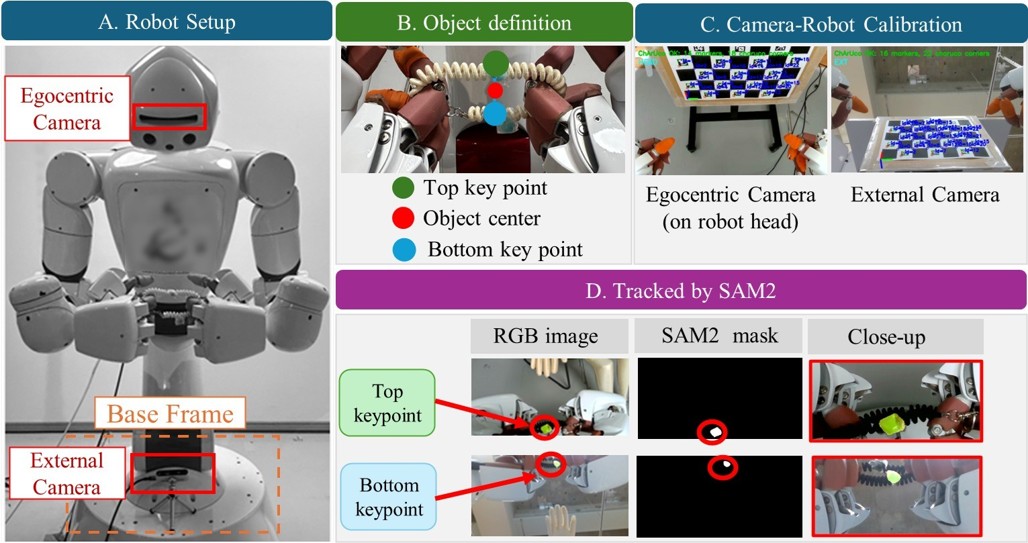}
    \caption{\textbf{Real-world setup.} A head-mounted egocentric camera estimates hand keypoints and the top opening marker, while an external camera observes the bottom marker. Both markers are tracked with SAM2. All views are transformed into the robot base frame.}
    \label{fig:real_world_setup}
\end{figure}

\textbf{Real world.} We verified that the same geometric observation interface can be instantiated on a bimanual humanoid platform (Fig.~\ref{fig:real_world_setup}). Human hand keypoints were estimated from the robot-mounted egocentric RGB camera using MediaPipe hand landmarks \cite{zhang2020mediapipehandsondevicerealtime}. Top and bottom (north/south) markers on the bracelet opening were tracked with SAM2 \cite{ravi2024sam2} rather than color segmentation; the object center is the midpoint of these markers. Because the bottom marker can be occluded from the head camera during insertion, an external Intel RealSense D435i observed the lower side of the bracelet. Both cameras were calibrated to the robot base frame with ChArUco boards, and the detected 3D keypoints were transformed into this common frame and fed to the learned policy in closed loop. The policy commanded the same $17$ arm and torso joints as in simulation; the head and grasping fingers remained fixed, and the bracelet was pre-grasped as in training. Trials used a static mannequin on a $20$-pose set in the training distribution. For hardware safety, we reduced the motion speed by approximately half during real-world deployment. A trial was successful when the bracelet passed over the hand and remained around the wrist.

\section{Results}
\label{sec:result}

\subsection{Overall Performance}
We evaluate success with the geometric containment test of Section~\ref{sec:citations}.
Each method is trained with three seeds. For each seed we select the checkpoint with the highest mean return on the evaluation environments over the $200$M-step run, and then score it on $50$ episodes independently sampled from the training pose distribution.
Fig.~\ref{fig:success_rate} shows the online success rate over the shared $200$M-step budget; Table~\ref{tab:ablation} reports the separate $50$-episode scores.

\begin{figure}[t]
    \centering
    \includegraphics[width=\linewidth]{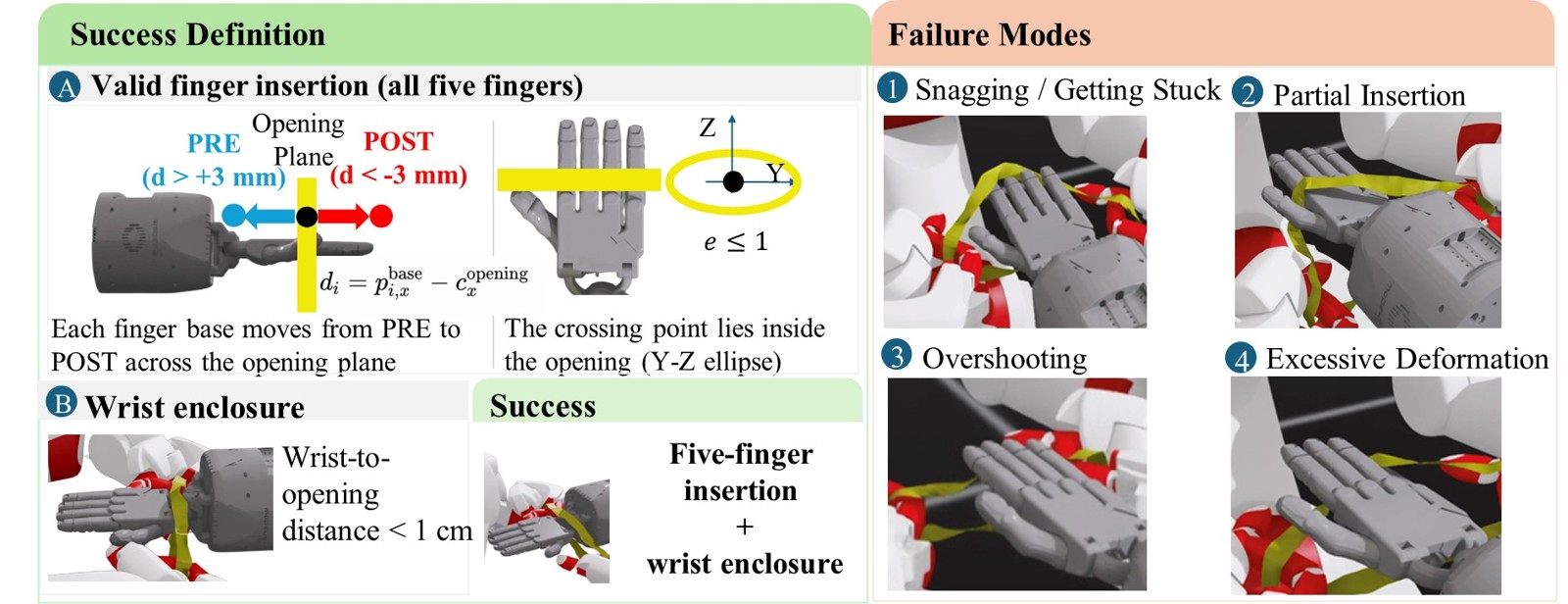}
    \caption{\textbf{Success criterion and failure modes.}
Left: success requires all five finger bases to cross the opening plane within the ellipse and the wrist-to-opening distance to be $<1\,\mathrm{cm}$.
Right: typical failures: snagging, partial insertion, overshooting, and excessive deformation.}
    \label{fig:failures}
\end{figure}

On the primary \emph{Deformable Bracelet} task, Full (soft gate + curriculum) finishes at $74.7\%$ and remains the top curve.
A hard gate + curriculum reaches $74.6\%$; a soft gate without curriculum reaches $71.1\%$.
Removing the insertion gate drops performance to $54.8\%$ (curriculum, no gate) and $41.2\%$ (no gate, no curriculum), and those bands do not meet the gated methods.
The gate is therefore what makes deformable insertion succeed: it stops the wrist term from pulling the opening onto incomplete fingers, which is the collapse and snag pattern in Fig.~\ref{fig:failures}.
Without the gate, the curriculum can activate the wrist reward before reliable insertion, leading to lower success.

The \emph{Rigid Bracelet} serves as a diagnostic control without opening collapse. Full finishes at $90.6\%$, followed by hard gate + curriculum at $84.8\%$ and soft gate without curriculum at $83.9\%$, while the no-gate variants remain lower at $71.6\%$ with curriculum and $53.3\%$ without.

\subsection{Ablation Studies}
\begin{figure}[t]
    \centering
    \includegraphics[width=0.83\linewidth]{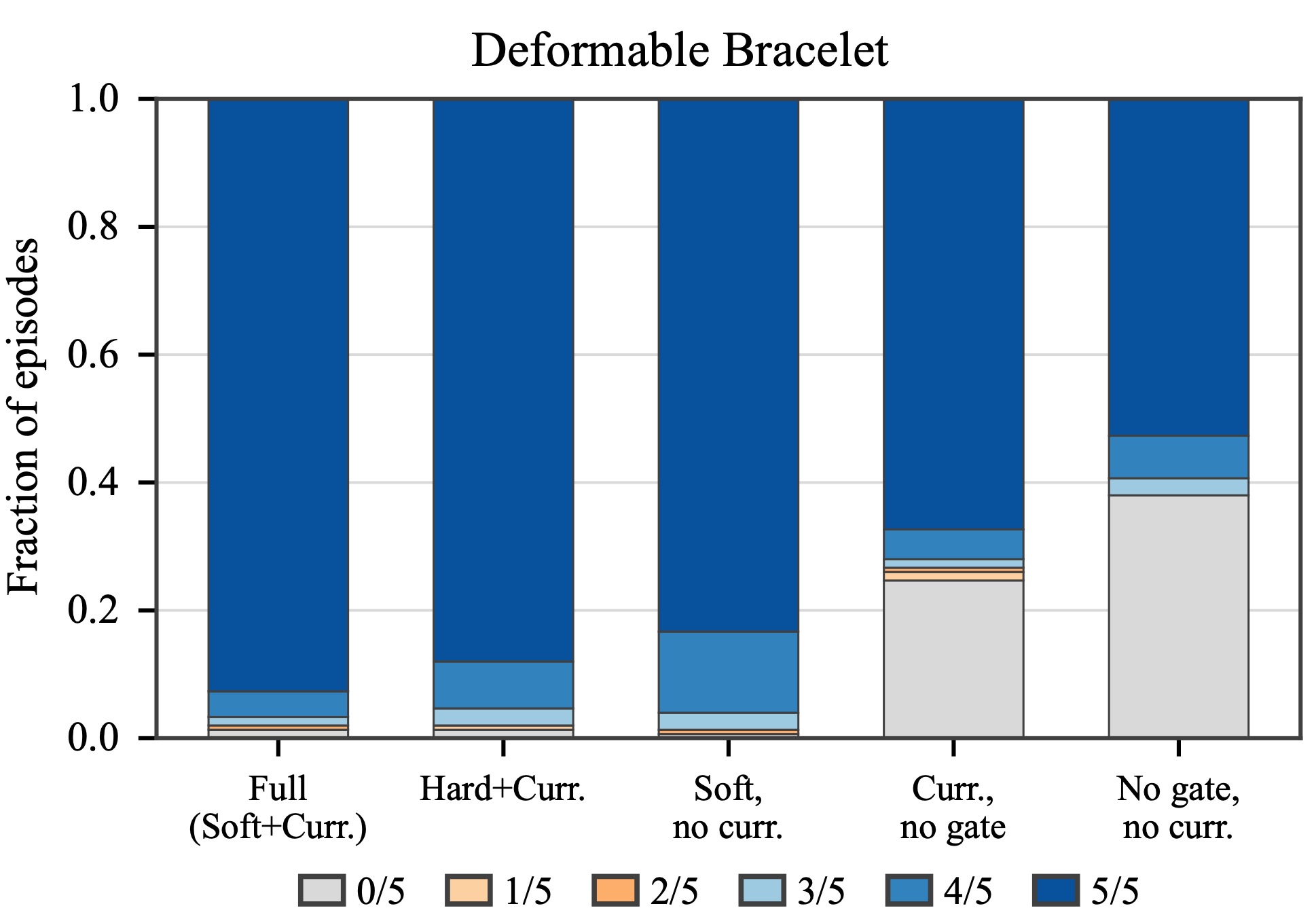}
    \caption{\textbf{Inserted-finger count at the final step.}
    \emph{Deformable Bracelet} post-training evaluation; three-seed mean over 50 episodes.}
    \label{fig:finger_hist}
\end{figure}

Table~\ref{tab:ablation} reports the same five reward variants in the post-training evaluation: three seeds, $50$ episodes per seed, without pooling episodes across seeds.
For all ablations, the poses were randomized by $\pm 5\,\mathrm{cm}$ in position and $\pm 10^{\circ}$ in roll, pitch, and yaw.
Evaluation episodes were independently sampled from the same pose-randomization distribution used during training under the same seed.
Fig.~\ref{fig:success_rate} shows only the ranking during training.
On the deformable bracelet, Full achieves $84.7\pm1.2\%$ success.
Removing the gate reduced success to $60.7\pm7.0\%$ with the curriculum
and to $48.0\pm9.2\%$ when both components were removed.
Replacing the soft gate with a hard gate yielded $76.7\pm5.0\%$, while
using the soft gate without the curriculum yields $71.3\pm9.9\%$.
Under Full, $92.7\%$ achieved five-finger insertion, but only $84.7\%$
met the full success criterion (Fig.~\ref{fig:finger_hist}).
Their mean minimum wrist-to-opening distance was $2.0\,\mathrm{cm}$ for
failures versus $0.8\,\mathrm{cm}$ for successes.
Thus, five-finger insertion alone is insufficient: success additionally
requires wrist enclosure, defined by a wrist-to-opening distance below
$1\,\mathrm{cm}$.

On the rigid diagnostic, Full achieves $90.0\pm7.2\%$ success, while
curriculum without a gate reaches $88.0\pm5.3\%$.
The hard-gate and soft-gate-without-curriculum variants achieve
$81.3\pm3.1\%$ and $77.3\pm15.0\%$, respectively, while removing both
components reduces success to $60.0\pm12.5\%$
(Table~\ref{tab:ablation}).

\begin{table}[t]
\centering
\caption{Post-training ablation evaluation of each seed's best checkpoint (highest mean evaluation return). \\Mean $\pm$ std over three seeds; $50$ episodes per seed.}
\label{tab:ablation}
\scriptsize
\setlength{\tabcolsep}{3pt}
\begin{tabularx}{\linewidth}{@{}l>{\centering\arraybackslash}X>{\centering\arraybackslash}X>{\centering\arraybackslash}X@{}}
\toprule
Method & Success (\%) & Fingers & Deflection\ ($^{\circ}$) \\
\midrule
\multicolumn{4}{c}{\emph{Deformable Bracelet}} \\
\textbf{Full (Soft Gate + Curriculum)}  & $\boldsymbol{84.7\pm 1.2}$& $4.85\pm 0.11$ & $0.74\pm 0.18$ \\
Hard Gate + Curriculum & $76.7\pm 5.0$ & $4.78\pm 0.05$ & $0.74\pm 0.10$ \\
Soft Gate + No Curriculum & $71.3\pm 9.9$ & $4.77\pm 0.08$ & $0.57\pm 0.15$ \\
Curriculum + No Gate & $60.7\pm 7.0$ & $3.62\pm 0.38$ & $2.30\pm 0.35$ \\
No Gate + No Curriculum & $48.0\pm 9.2$ & $2.98\pm 0.24$ & $1.63\pm 1.10$ \\
\midrule
\multicolumn{4}{c}{\emph{Rigid Bracelet} (diagnostic)} \\
\textbf{Full (Soft Gate + Curriculum)} & $\boldsymbol{90.0\pm 7.2}$& $4.87\pm 0.11$ & $1.44\pm 0.90$ \\
Hard Gate + Curriculum & $81.3\pm 3.1$ & $4.63\pm 0.22$ & $1.76\pm 0.75$ \\
Soft Gate + No Curriculum & $77.3\pm 15.0$ & $4.77\pm 0.02$ & $0.98\pm 0.35$ \\
Curriculum + No Gate & $88.0\pm 5.3$ & $4.65\pm 0.17$ & $1.80\pm 0.99$ \\
No Gate + No Curriculum & $60.0\pm 12.5$ & $3.44\pm 0.75$ & $2.86\pm 1.12$ \\
\bottomrule
\end{tabularx}
\end{table}

\subsection{Pose Robustness}
Table~\ref{tab:robustness} reports a separate axis-wise pose-robustness
evaluation using policies trained and evaluated without wrist roll;
these are not the checkpoints evaluated in Table~\ref{tab:ablation}.
Position-only shifts of $\pm 2.5$, $\pm 5.0$, and $\pm 7.5\,\mathrm{cm}$ retain $88\%$, $92\%$, and $80\%$ success. Joint yaw and pitch of $\pm 5^{\circ}$, $\pm 10^{\circ}$, and $\pm 15^{\circ}$ retain $98\%$, $92\%$, and $78\%$. Combining translation with the same yaw--pitch ranges remains usable at the smallest setting ($47/50$, $94\%$), but falls to $33/50$ ($66\%$) and $26/50$ ($52\%$) as both grow. We do not interpret these results as evidence of broad robustness: moderate translation-only or orientation-only changes are handled, and large simultaneous changes are not. The typical failures are a thumb-side snag, a collapsed opening, a yaw--pitch offset that breaks the little-to-thumb order, and the same defects under a combined pose offset (Fig.~\ref{fig:failures}). Finger insertion often stays above four fingers even when the episode is a failure, so partial progress is not counted as success. Finger-joint deflection increases with the combined range; it is a kinematic contact proxy, not a force or safety measurement. A yaw--pitch error rotates the little-to-thumb order relative to the opening, so the gate saturates later and snagging becomes more likely.

\begin{table}[t]
\vspace{0.2cm}
\centering
\caption{Pose robustness on the \emph{Deformable Bracelet} task.\\ Success is reported as count$/50$ episodes.}
\label{tab:robustness}
\scriptsize
\setlength{\tabcolsep}{2.5pt}
\begin{tabularx}{\linewidth}{@{}l>{\raggedright\arraybackslash}Xccc@{}}
\toprule
Condition & Range & Success & Fingers & Deflection\ ($^{\circ}$) \\
\midrule
Pos.
& $\pm 2.5\,\mathrm{cm}$ & $44/50$ & $4.92\pm 0.40$ & $0.44\pm 0.29$ \\
& $\pm 5.0\,\mathrm{cm}$ & $46/50$ & $4.98\pm 0.14$ & $0.47\pm 0.31$ \\
& $\pm 7.5\,\mathrm{cm}$ & $40/50$ & $4.92\pm 0.44$ & $0.76\pm 0.51$ \\
\midrule
Yaw/Pitch
& $\pm 5^{\circ}$ & $49/50$ & $4.98\pm 0.14$ & $0.34\pm 0.47$ \\
& $\pm 10^{\circ}$ & $46/50$ & $4.76\pm 0.74$ & $0.60\pm 1.24$ \\
& $\pm 15^{\circ}$ & $39/50$ & $4.82\pm 0.83$ & $1.06\pm 1.20$ \\
\midrule
Pos.+Yaw/Pitch
& $\pm 2.5\,\mathrm{cm}$, $\pm 5^{\circ}$ & $47/50$ & $4.94\pm 0.24$ & $0.74\pm 1.05$ \\
& $\pm 5.0\,\mathrm{cm}$, $\pm 10^{\circ}$ & $33/50$ & $4.42\pm 1.05$ & $0.90\pm 1.39$ \\
& $\pm 7.5\,\mathrm{cm}$, $\pm 15^{\circ}$ & $26/50$ & $4.36\pm 1.06$ & $1.41\pm 1.74$ \\
\bottomrule
\end{tabularx}
\end{table}

\subsection{Real-World Policy Evaluation}
Each simulation-trained policy is executed without additional training on this $20$-pose mannequin set: $2$ nominal poses, $\pm 3\,\mathrm{cm}$ along $x/y/z$, $\pm 10^{\circ}$ roll/pitch/yaw, and $6$ combined perturbations (Table~\ref{tab:real_eval}).
The full method succeeds on $15/20$ trials ($75\%$).
The remaining methods trail: hard gate + curriculum $12/20$ ($60\%$), a soft gate without curriculum $10/20$ ($50\%$), curriculum without a gate $9/20$ ($45\%$), and neither $6/20$ ($30\%$).
Residual failures include overshoot, snagging, and opening twist that geometry does not encode.
Overshoot is the main remaining failure mode. The benefit of the gate is most evident for combined perturbations and negative orientations, while translation-only perturbations show little difference between methods.
\begin{table}[t]
    \vspace{0.1cm}

\centering
\caption{Real-robot success across $20$ mannequin poses, with one closed-loop trial per pose.}
\label{tab:real_eval}
\scriptsize
\setlength{\tabcolsep}{4pt}
\begin{tabular}{@{}lc@{}}
\toprule
Method & Success \\
\midrule
Full (Soft Gate + Curriculum) & $15/20$ ($75\%$) \\
Hard Gate + Curriculum & $12/20$ ($60\%$) \\
Soft Gate + No Curriculum & $10/20$ ($50\%$) \\
Curriculum + No Gate & $9/20$ ($45\%$) \\
No Gate + No Curriculum & $6/20$ ($30\%$) \\
\bottomrule
\end{tabular}
\end{table}

\section{Discussion}
\label{sec:discussion}

Our results show that bracelet insertion can be learned from compact
relative hand--object geometry without representing the full deformable
object state.
Success requires a sequence of valid finger insertions followed by wrist
enclosure, rather than proximity to the wrist alone
(Fig.~\ref{fig:rew_set}).
The ablations further indicate that the insertion gate provides the
largest improvement for the deformable bracelet
(Table~\ref{tab:ablation}), while the curriculum provides an additional
gain when combined with the gate.
Without gating, both curriculum variants remain substantially below the
gated methods.
This difference becomes smaller for the rigid diagnostic: Full achieves
$90.0\%$, while curriculum without a gate reaches $88.0\%$.
This is consistent with the gate being particularly useful when insertion
must be maintained despite deformation of the opening, although the
remaining collapse, snag, overshoot, and incomplete-enclosure failures
show that gating does not eliminate these interactions
(Fig.~\ref{fig:failures}).

Related failure modes also appear on the real robot.
Residual failures include overshoot, snagging, and opening twist
(Table~\ref{tab:real_eval}), illustrating the sensitivity of
close-fitting insertion to relative pose and contact.
A small pose error, or one finger contacting the rim before the others,
can alter the subsequent insertion sequence.
The robustness experiments similarly show successful transfer under
moderate translation-only and joint yaw--pitch variations, with
performance degrading under larger combined perturbations
(Table~\ref{tab:robustness}).

The simulation-to-real difference is not captured by the success-rate
alone.
Simulation provides exact hand and bracelet keypoints, whereas
the real system relies on geometry estimated from visual observations.
Perception and calibration errors therefore introduce uncertainty into
the relative geometry observed by the policy.
Moreover, friction, opening tension, and hand compliance are not fully
captured by the simulator, so similar geometric observations and commands
can produce different contact outcomes.
The recurrence of overshoot, snagging, and deformation-related failures
in both domains suggests that the underlying insertion challenge is
shared, while perception error and unmodeled contact dynamics contribute
additional sources of error in the real system.
Given only $20$ trials per method, we do not interpret differences among
the intermediate ablations as evidence of a changed mechanism; rather,
the real-robot results establish transfer of the learned policy and expose
the remaining sensitivity to perception and contact.

A key limitation is that the current formulation evaluates insertion
geometrically and does not explicitly account for contact forces.
Extending the policy with force-aware observations and objectives will
be important for physical human assistance.
Our experiments also focus on the insertion stage with a pre-grasped
bracelet and a stationary target hand.
Extending the setting to diverse hand geometries, human motion, and
contact conditions is an important direction toward more general
close-fitting dressing assistance.

\section{CONCLUSIONS}
We presented a geometry-conditioned RL framework for close-fitting wearable dressing, formulated as articulated insertion through a narrow, deformable opening.
A finger-wise insertion gate and a staged finger-to-wrist curriculum enable coordinated insertion while accounting for the geometric constraints imposed by the wearable opening.
On the deformable bracelet, the full method achieves $84.7\pm1.2\%$ post-training success, with ablations showing degraded performance when the gating or curriculum structure is altered or removed.
The rigid diagnostic further isolates geometric insertion from failures caused by opening deformation, while real-robot experiments demonstrate that the same geometric representation can be reconstructed from vision and used for closed-loop control, achieving $15/20$ successful mannequin trials.
Together, these results support task-relevant relative geometry as a compact representation for learning contact-rich insertion into close-fitting deformable wearables.

Future work will incorporate force-aware observations, broaden the training distribution, and evaluate physical interaction with real users.
Beyond insertion, an important next step is to extend the learned behavior to subsequent garment manipulation, including pulling the wearable along the body part toward its final dressed configuration.

\section*{ACKNOWLEDGMENT}
Cursor was used during this work to assist with code development, create preliminary prototypes for figures, and refine the grammar and phrasing of the manuscript.

\bibliographystyle{ieeetr}
\bibliography{root}

@article{wang2025dexgarmentlab,
   title={DexGarmentLab: Dexterous Garment Manipulation Environment with Generalizable Policy},
   author={Wang, Yuran and Wu, Ruihai and Chen, Yue and Wang, Jiarui and Liang, Jiaqi and Zhu, Ziyu and Geng, Haoran and Malik, Jitendra and Abbeel, Pieter and Dong, Hao},
   journal={Advances in Neural Information Processing Systems},
   year={2025},
}

@inproceedings{hoang2025geometryaware,
  title={Geometry-aware {RL} for Manipulation of Varying Shapes and Deformable Objects},
  author={Tai Hoang and Huy Le and Philipp Becker and Vien Anh Ngo and Gerhard Neumann},
  booktitle={The Thirteenth International Conference on Learning Representations},
  year={2025},
  url={https://openreview.net/forum?id=7BLXhmWvwF}
}

@inproceedings{tamei2011reinforcement,
  title={Reinforcement learning of clothing assistance with a dual-arm robot},
  author={Tamei, Tomoya and Matsubara, Takamitsu and Rai, Akshara and Shibata, Tomohiro},
  booktitle={2011 11th IEEE-RAS International Conference on Humanoid Robots},
  pages={733--738},
  year={2011},
  organization={IEEE},
  doi={10.1109/Humanoids.2011.6100915}
}

@article{zhang2022learning,
  title={Learning garment manipulation policies toward robot-assisted dressing},
  author={Zhang, Fan and Demiris, Yiannis},
  journal={Science robotics},
  volume={7},
  number={65},
  pages={eabm6010},
  year={2022},
  publisher={American Association for the Advancement of Science},
  doi = {10.1126/scirobotics.abm6010}
}

@ARTICLE{zhang2023visual,
  author={Zhang, Fan and Demiris, Yiannis},
  journal={IEEE Robotics and Automation Letters}, 
  title={Visual-Tactile Learning of Garment Unfolding for Robot-Assisted Dressing}, 
  year={2023},
  volume={8},
  number={9},
  pages={5512-5519},
  doi={10.1109/LRA.2023.3296371}}

@article{sun2024force,
  title={Force-Constrained Visual Policy: Safe Robot-Assisted Dressing via Multi-Modal Sensing},
  author={Sun, Zhanyi and Wang, Yufei and Held, David and Erickson, Zackory},
  journal={IEEE Robotics and Automation Letters},
  year={2024},
  volume={9},
  pages={4178--4185},
  publisher={IEEE},
  doi={10.1109/LRA.2024.3375712}
}

@article{kotsovolis2024garment,
 author={Kotsovolis, Stelios and Demiris, Yiannis},
  journal={IEEE Robotics and Automation Letters}, 
  title={Garment Diffusion Models for Robot-Assisted Dressing}, 
  year={2025},
  volume={10},
  number={2},
  pages={1217-1224},
  doi={10.1109/LRA.2024.3518104}}

@article{wang2023one, 
    AUTHOR    = {Yufei Wang AND Zhanyi Sun AND Zackory Erickson AND David Held}, 
    TITLE     = {{One Policy to Dress Them All: Learning to Dress People with Diverse Poses and Garments}}, 
    journal = {Proceedings of Robotics: Science and Systems, Daegu, Republic of Korea}, 
    YEAR      = {2023}, 
    ADDRESS   = {Daegu, Republic of Korea}, 
    MONTH     = {July}, 
    DOI       = {10.15607/RSS.2023.XIX.008} 
}

@article{chen2026feeding,
  title={Feeding, grooming, dressing, and body repositioning: categorizing four pillars of learning-based manipulation for robotic caregiving},
  author={Chen, Wenkai and Huang, Shifeng and Zhang, Hui and Tu, Yuyang and Lu, Yao and Li, Qingdu and Zhang, Jianwei},
  journal={Artificial Intelligence Review},
  volume={59},
  number={4},
  pages={113},
  year={2026},
  publisher={Springer},
  doi = {https://doi.org/10.1007/s10462-026-11524-7}
}

@article{pignat2017learning,
  title={Learning adaptive dressing assistance from human demonstration},
  author={Pignat, Emmanuel and Calinon, Sylvain},
  journal={Robotics and Autonomous Systems},
  volume={93},
  pages={61--75},
  year={2017},
  publisher={Elsevier},
  doi={https://doi.org/10.1016/j.robot.2017.03.017}
}

@ARTICLE{Clegg2020learning,
  author={Clegg, Alexander and Erickson, Zackory and Grady, Patrick and Turk, Greg and Kemp, Charles C. and Liu, C. Karen},
  journal={IEEE Robotics and Automation Letters}, 
  title={Learning to Collaborate From Simulation for Robot-Assisted Dressing}, 
  year={2020},
  volume={5},
  number={2},
  pages={2746-2753},
  doi={10.1109/LRA.2020.2972852}}

@INPROCEEDINGS{chen2025graphgarment,
  author={Chen, Wei and Li, Kelin and Lee, Dongmyoung and Chen, Xiaoshuai and Zong, Rui and Kormushev, Petar},
  booktitle={2025 IEEE/RSJ International Conference on Intelligent Robots and Systems (IROS)}, 
  title={GraphGarment: Learning Garment Dynamics for Bimanual Cloth Manipulation Tasks}, 
  year={2025},
  volume={},
  number={},
  pages={7615-7621},
  doi={10.1109/IROS60139.2025.11246283}}

@article{joshi2019framework,
  title={A framework for robotic clothing assistance by imitation learning},
  author={Joshi, Ravi Prakash and Koganti, Nishanth and Shibata, Tomohiro},
  journal={Advanced Robotics},
  volume={33},
  number={22},
  pages={1156--1174},
  year={2019},
  publisher={Taylor \& Francis},
  doi={https://doi.org/10.1080/01691864.2019.1636715}
}

@ARTICLE{Tsukakoshi2026close,
  author={Tsukakoshi, Takuma and Miyake, Tamon and Ogata, Tetsuya and Wang, Yushi and Akaishi, Takumi and Sugano, Shigeki},
  journal={IEEE Robotics and Automation Letters}, 
  title={Close-Fitting Dressing Assistance Based on State Estimation of Feet and Garments With Semantic-Based Visual Attention}, 
  year={2026},
  volume={11},
  number={4},
  pages={3923-3930},
  doi={10.1109/LRA.2026.3664535}}

@article{schulman_proximal_2017,
  title={Proximal policy optimization algorithms},
  author={Schulman, John and Wolski, Filip and Dhariwal, Prafulla and Radford, Alec and Klimov, Oleg},
  journal={arXiv preprint arXiv:1707.06347},
  year={2017}
}

@inproceedings{miller2025tactilerl,
  author    = {Miller, Elle and McInroe, Trevor and Abel, David and Mac Aodha, Oisin and Vijayakumar, Sethu},
  title     = {Enhancing Tactile-based Reinforcement Learning for Robotic Control},
  booktitle   = {NeurIPS},
  year      = {2025},
}

@article{isaaclab,
   author={Mittal, Mayank and Yu, Calvin and Yu, Qinxi and Liu, Jingzhou and Rudin, Nikita and Hoeller, David and Yuan, Jia Lin and Singh, Ritvik and Guo, Yunrong and Mazhar, Hammad and Mandlekar, Ajay and Babich, Buck and State, Gavriel and Hutter, Marco and Garg, Animesh},
   journal={IEEE Robotics and Automation Letters},
   title={Orbit: A Unified Simulation Framework for Interactive Robot Learning Environments},
   year={2023},
   volume={8},
   number={6},
   pages={3740-3747},
   doi={10.1109/LRA.2023.3270034}
}

@article{serrano2023skrl,
  author  = {Antonio Serrano-Muñoz and Dimitrios Chrysostomou and Simon Bøgh and Nestor Arana-Arexolaleiba},
  title   = {skrl: Modular and Flexible Library for Reinforcement Learning},
  journal = {Journal of Machine Learning Research},
  year    = {2023},
  volume  = {24},
  number  = {254},
  pages   = {1--9},
  url     = {http://jmlr.org/papers/v24/23-0112.html}
}

@article{hao2025force,
  title={Force-Modulated Visual Policy for Robot-Assisted Dressing with Arm Motions},
  author={Hao, Alexis Yihong and Wang, Yufei and Ravie, Navin Sriram and Hegde, Bharath and Held, David and Erickson, Zackory},
  journal={Conference on Robot Learning (CoRL)},
  year={2026},
  pages = {5483-5505},
  url ={https://proceedings.mlr.press/v305/hao25b.html}
}

@inproceedings{lin2021softgym,
  title={Softgym: Benchmarking deep reinforcement learning for deformable object manipulation},
  author={Lin, Xingyu and Wang, Yufei and Olkin, Jake and Held, David},
  booktitle={Conference on Robot Learning},
  pages={432--448},
  year={2021},
  organization={PMLR}
}

@article{puthuveetil2022bodies,
  title={Bodies uncovered: Learning to manipulate real blankets around people via physics simulations},
  author={Puthuveetil, Kavya and Kemp, Charles C and Erickson, Zackory},
  journal={IEEE Robotics and Automation Letters},
  volume={7},
  number={2},
  pages={1984--1991},
  year={2022},
  publisher={IEEE},
  doi={https://doi.org/10.1109/LRA.2022.3142732}
}

@article{hoque2020visuospatial,
  title={Visuospatial foresight for multi-step, multi-task fabric manipulation},
  author={Hoque, Ryan and Seita, Daniel and Balakrishna, Ashwin and Ganapathi, Aditya and Tanwani, Ajay Kumar and Jamali, Nawid and Yamane, Katsu and Iba, Soshi and Goldberg, Ken},
  journal={Robotics: Science and Systems},
  year={2020}
}

@inproceedings{erickson2020assistive,
  title={Assistive gym: A physics simulation framework for assistive robotics},
  author={Erickson, Zackory and Gangaram, Vamsee and Kapusta, Ariel and Liu, C Karen and Kemp, Charles C},
  booktitle={2020 IEEE International Conference on Robotics and Automation (ICRA)},
  pages={10169--10176},
  year={2020},
  organization={IEEE}
}

@article{tsurumine2019deep,
  title={Deep reinforcement learning with smooth policy update: Application to robotic cloth manipulation},
  author={Tsurumine, Yoshihisa and Cui, Yunduan and Uchibe, Eiji and Matsubara, Takamitsu},
  journal={Robotics and Autonomous Systems},
  volume={112},
  pages={72--83},
  year={2019},
  publisher={Elsevier}
}

@article{louis2026se3,
title = {SE(3)-based representation of object deformations for robotic shape servoing: Formulation and assessment},
journal = {Robotics and Autonomous Systems},
volume = {200},
pages = {105403},
year = {2026},
issn = {0921-8890},
doi = {https://doi.org/10.1016/j.robot.2026.105403},
author = {Louis Dehaybe and Olivier Brüls},
}

@article{wang2025goal,
  title={Goal-Conditioned Model Simplification for 1-D and 2-D Deformable Object Manipulation},
  author={Wang, Shengyin and Leonetti, Matteo and Dogar, Mehmet},
  journal={IEEE Transactions on Robotics},
  year={2025},
  publisher={IEEE},
  doi={https://doi.org/10.1109/TRO.2025.3577052}
}

@misc{ravi2024sam2,
      title={SAM 2: Segment Anything in Images and Videos},
      author={Nikhila Ravi and Valentin Gabeur and Yuan-Ting Hu and Ronghang Hu and Chaitanya Ryali and Tengyu Ma and Haitham Khedr and Roman R{\"a}dle and Chloe Rolland and Laura Gustafson and Eric Mintun and Junting Pan and Kalyan Vasudev Alwala and Nicolas Carion and Chao-Yuan Wu and Ross Girshick and Piotr Doll{\'a}r and Christoph Feichtenhofer},
      year={2024},
      eprint={2408.00714},
      archivePrefix={arXiv},
      primaryClass={cs.CV},
      url={https://arxiv.org/abs/2408.00714},
}

@misc{zhang2020mediapipehandsondevicerealtime,
      title={MediaPipe Hands: On-device Real-time Hand Tracking}, 
      author={Fan Zhang and Valentin Bazarevsky and Andrey Vakunov and Andrei Tkachenka and George Sung and Chuo-Ling Chang and Matthias Grundmann},
      year={2020},
      eprint={2006.10214},
      archivePrefix={arXiv},
      primaryClass={cs.CV},
      url={https://arxiv.org/abs/2006.10214}, 
}

@article{zhu2024you,
  title={Do you need a hand?--a bimanual robotic dressing assistance scheme},
  author={Zhu, Jihong and Gienger, Michael and Franzese, Giovanni and Kober, Jens},
  journal={IEEE Transactions on Robotics},
  volume={40},
  pages={1906--1919},
  year={2024},
  publisher={IEEE},
  doi={https://doi.org/10.1109/TRO.2024.3366008}
}

@article{wang2025robot,
  title={Robot Deformable Object Manipulation via NMPC-Generated Demonstrations in Deep Reinforcement Learning},
  author={Wang, Haoyuan and Dong, Zihao and Zhu, Tong and Lei, Hongliang and Shi, Weizhuang and Zhang, Zejia and Luo, Wei and Wan, Weiwei and Chen, Xinxing and Huang, Jian},
  journal={IEEE Transactions on Automation Science and Engineering},
  volume={22},
  pages={23566--23578},
  year={2025},
  publisher={IEEE},
  doi={https://doi.org/10.1109/TASE.2025.3627775}
}

@inproceedings{ye2022rcareworld,
  title={{RCareWorld}: A Human-centric Simulation World for Caregiving Robots},
  author={Ye, Ruolin and Xu, Wenqiang and Fu, Haoyuan and Lai, Yapeng Kevin and Liu, C. Karen and Wang, He},
  booktitle={2022 IEEE/RSJ International Conference on Intelligent Robots and Systems (IROS)},
  pages={33--40},
  year={2022},
  organization={IEEE}
}

@article{longhini2025unfolding,
  title={Unfolding the literature: A review of robotic cloth manipulation},
  author={Longhini, Alberta and Wang, Yufei and Garcia-Camacho, Irene and Blanco-Mulero, David and Moletta, Marco and Welle, Michael and Aleny{\`a}, Guillem and Yin, Hang and Erickson, Zackory and Held, David and others},
  journal={Annual Review of Control, Robotics, and Autonomous Systems},
  volume={8},
  number={1},
  pages={295--322},
  year={2025},
  publisher={Annual Reviews},
  doi={https://doi.org/10.1146/annurev-control-022723-033252}
}

@inproceedings{liang2025openrobocare,
  title={OpenRoboCare: A Multimodal Multi-Task Expert Demonstration Dataset for Robot Caregiving},
  author={Liang, Xiaoyu and Liu, Ziang and Lin, Kelvin and Gu, Edward and Ye, Ruolin and Nguyen, Tam and Hsu, Cynthia and Wu, Zhanxin and Yang, Xiaoman and Cheung, Christy Sum Yu and others},
  booktitle={2025 IEEE/RSJ International Conference on Intelligent Robots and Systems (IROS)},
  pages={2661--2668},
  year={2025},
  organization={IEEE},
  doi={https://doi.org/10.1109/IROS60139.2025.11247516}
}

\end{document}